\documentclass{article}

\usepackage[preprint]{neurips_2025}

\usepackage[utf8]{inputenc} 
\usepackage[T1]{fontenc}    
\usepackage{hyperref}       
\usepackage{url}            
\usepackage{booktabs}       
\usepackage{amsfonts}       
\usepackage{dsfont}         
\usepackage{nicefrac}       
\usepackage{microtype}      
\usepackage{xcolor}         

\usepackage{mymathdef}

\renewcommand{\cite}[1]{\citep{#1}}

\title{SoftCoT++: Test-Time Scaling with Soft Chain-of-Thought Reasoning}

\author{
Yige Xu${^{1,3}}$\thanks{\ \  The first two authors contributed equally.}, Xu Guo$^{2,4}$\footnotemark[1]\;\thanks{\ \  Corresponding author.}, Zhiwei Zeng$^2$, Chunyan Miao$^{1,2,3}$ \\
$^1$Joint NTU-UBC Research Centre of Excellence in Active Living for the Elderly \\
$^2$Alibaba-NTU Global e-Sustainability CorpLab (ANGEL)\\
$^3$College of Computing and Data Science, Nanyang Technological University, Singapore \\
$^4$KTH Royal Institute of Technology, Sweden\\
\texttt{\{yige002,xu008\}@e.ntu.edu.sg}, \texttt{\{zhiwei.zeng,ascymiao\}@ntu.edu.sg}
}

\begin{document}

\maketitle

\begin{abstract}
Test-Time Scaling (TTS) refers to approaches that improve reasoning performance by allocating extra computation during inference, without altering the model's parameters.
While existing TTS methods operate in a discrete token space by generating more intermediate steps, recent studies in Coconut and SoftCoT have demonstrated that thinking in the continuous latent space can further enhance the reasoning performance. Such latent thoughts encode informative thinking without the information loss associated with autoregressive token generation, sparking increased interest in continuous-space reasoning.
Unlike discrete decoding, where repeated sampling enables exploring diverse reasoning paths, latent representations in continuous space are fixed for a given input, which limits diverse exploration, as all decoded paths originate from the same latent thought.
To overcome this limitation, we introduce {\bf SoftCoT++} to extend SoftCoT to the Test-Time Scaling paradigm by enabling diverse exploration of thinking paths.
Specifically, we perturb latent thoughts via multiple specialized initial tokens and apply contrastive learning to promote diversity among soft thought representations.
Experiments across five reasoning benchmarks and two distinct LLM architectures demonstrate that SoftCoT++ significantly boosts SoftCoT and also outperforms SoftCoT with self-consistency scaling. Moreover, it shows strong compatibility with conventional scaling techniques such as self-consistency. Source code is available at \url{https://github.com/xuyige/SoftCoT}.
\end{abstract}

\section{Introduction}

In recent years, performance improvements in Large Language Models (LLMs)~\cite{DBLP:conf/nips/BrownMRSKDNSSAA20,chowdhery2023palm,openai2023gpt4,dubey2024llama,DBLP:journals/corr/abs-2412-15115,deepseek2025deepseekr1,qwen3} have largely stemmed from scaling up training-time compute.
These large-scale models exhibit emergent reasoning abilities, notably through Chain-of-Thought (CoT) prompting~\cite{DBLP:conf/nips/Wei0SBIXCLZ22}, which generates explicit intermediate steps to enhance answer accuracy.
Building on this foundation, a new scaling paradigm known as Test-Time Scaling (TTS)~\cite{DBLP:conf/iclr/0002WSLCNCZ23,snell2024scaling,brown2024large,muennighoff2025s1} has emerged, aiming to further enhance reasoning performance by allocating additional computation at inference time, without modifying the model parameters.

Existing TTS methods can be broadly classified into two regimes: parallel scaling~\cite{DBLP:conf/iclr/0002WSLCNCZ23,renze-2024-effect,brown2024large}, and sequential scaling~\cite{DBLP:conf/nips/MadaanTGHGW0DPY23,snell2024scaling,chen2025iterative}. Parallel scaling, such as Best-of-N (BoN)~\cite{BoN} and Self-Consistency (SC)~\cite{DBLP:conf/iclr/0002WSLCNCZ23}, generates multiple reasoning chains via independent sampling and aggregates the final answer through a fusion mechanism. In contrast, sequential scaling directs later computations based explicitly on earlier intermediate steps~\cite{DBLP:conf/iclr/ZhouSHWS0SCBLC23}. In this work, we primarily focus on parallel scaling by encouraging the generation of diverse reasoning chains. Notably, both paradigms operate within the discrete token space for generating intermediate steps, potentially limiting their ability to capture nuanced or continuous reasoning dynamics.

Recently, the idea of reasoning in a continuous latent space has garnered increasing attention in the community~\cite{DBLP:journals/corr/abs-2412-06769,DBLP:journals/corr/abs-2412-13171,DBLP:journals/corr/abs-2501-19201}. Studies in Coconut~\cite{DBLP:journals/corr/abs-2412-06769} and SoftCoT~\cite{xu2025softcot} demonstrate that leveraging latent thoughts can enhance subsequent reasoning quality.
Notably, SoftCoT freezes the LLM and utilizes a fixed small assistant model to generate soft thoughts. It outperforms Coconut on recent LLMs from LLaMA and Qwen families, while Coconut even underperforms zero-shot CoT prompting, which already yields strong results with modern LLMs.
Nevertheless, scaling in the continuous space remains challenging, as it does not naturally support multi-path sampling as in classical TTS methods.

Unlike discrete-space reasoning, which naturally allows sampling multiple reasoning paths from a probability distribution $P(x_i\mid x_{<i})$ over tokens $x_i\in \mathcal{V}$, continuous-space reasoning outputs a deterministic latent thought for a given question. There is no explicit distribution for sampling diverse latent thoughts. To simulate stochastic sampling in continuous space, we conducted pilot experiments with SoftCoT by adding small perturbations to a single latent soft thought to approximate the stochasticity. We compare using diverse soft thoughts (SoftCoT-P) for parallel scaling with conventional discrete-space scaling via token sampling (SoftCoT-SC) and find that these two scaling strategies can achieve comparable performance. This confirms the feasibility of continuous-space scaling.
Additionally, theoretical analysis (in Appendix~\ref{appendix:self-consistency}) indicates that parallel scaling with majority voting helps only when the base LLM already reasons well. We therefore adopt SoftCoT as the foundation for parallel scaling in continuous space, given its strong performance on recent state-of-the-art LLMs.

This paper introduces {\bf SoftCoT++}, the first framework for scaling continuous-space CoT to enhance LLM reasoning performance. Building on SoftCoT, we split the generation process into a {\bf thinking} stage (latent soft thoughts) and a {\bf reasoning} stage (token generation). SoftCoT++ scales the latent thinking stage while remaining fully compatible with conventional token-level scaling during reasoning.
Specifically, we introduce multiple specialized initial tokens that serve as distinct prompts to the assistant model, prompting it to generate multiple soft thought representations for a given input. This design simulates parallel scaling in discrete space by generating multiple latent reasoning paths simultaneously.
To further promote exploring distinct reasoning paths, we employ a contrastive learning objective to explicitly push the soft thoughts apart in the latent space. Theoretical analysis (in Appendix~\ref{appendix:proof-of-lemma}) shows that SoftCoT++ can provide a better approximation to the true latent-thought distribution than random perturbation. Together, distinct initialization paired with contrastive learning enables SoftCoT++ to scale reasoning at test time while preserving the efficiency and stability benefits of continuous latent thinking.

Following SoftCoT~\cite{xu2025softcot}, we evaluate SoftCoT++ on five reasoning benchmarks and two state-of-the-art LLM architectures. The five benchmarks include mathematical reasoning, commonsense reasoning, and symbolic reasoning. The two LLM architectures include LLaMA-3.1 series~\cite{dubey2024llama} and Qwen3 series~\cite{qwen3}. Experimental results show that SoftCoT++ consistently outperforms all baselines, which apply test-time scaling in discrete token space, across architectures and tasks. This highlights the effectiveness of applying test-time scaling for continuous-space reasoning.
Since SoftCoT++ scales latent thoughts on the \textit{thinking} stage, while existing discrete-space scaling methods like SC scale on the \textit{reasoning} stage, the two mechanisms can be complementary. We demonstrate through experiments that SoftCoT++ combined with SC (Table~\ref{table:softcot++-experiment-self-consistency}) can amplify the overall scaling effect.

\section{Related Works}

\subsection{Test-Time Scaling}

Test-time scaling (TTS) has emerged as a pivotal strategy in enhancing the performance of LLMs by allocating additional computational resources during inference. This approach shifts the traditional emphasis from extensive pretraining to optimizing inference-time computation, enabling models to tackle complex tasks more effectively. Following \citet{muennighoff2025s1} and \citet{zhang2025and}, we classify test-time scaling methods into: (1) {\bf Parallel Scaling}~\cite{DBLP:conf/iclr/0002WSLCNCZ23,brown2024large,snell2024scaling,liu2025pearl}, where parallel computes multiple reasoning chains independently, (2) {\bf Sequential Scaling}~\cite{DBLP:conf/nips/MadaanTGHGW0DPY23,DBLP:conf/iclr/ChenLSZ24,muennighoff2025s1}, where computes a longer reasoning chain and generates the chain sequentially, and (3) {\bf Hybrid Scaling}~\cite{DBLP:conf/nips/YaoYZS00N23,gandhi2024stream,DBLP:conf/iclr/WangWAZZ25}, where combines the parallel scaling and sequential scaling methods. In this paper, we mainly focus on parallel test-time scaling, which can be adopt on large-scale LLMs efficiently.

As conclued by \citet{zhang2025and}, parallel scaling improves test-time performance by generating multiple reasoning chains in parallel, and then aggregating them together to the final answer. Early evidence that sampling multiple reasoning chains and voting improves robustness came from Self-Consistency (SC)~\cite{DBLP:conf/iclr/0002WSLCNCZ23}, inspiring subsequent studies on how many chains to sample for a fixed compute envelope~\cite{snell2024scaling}. \citet{li2025s} suggest that the chance of finding the correct answer improves while increasing the number of generated responses, which is empirically summarized by a log-linear scaling law~\cite{brown2024large}. Despite the effectiveness of these approaches, the majority of existing parallel test-time scaling methods rely on discrete token-by-token generation, which imposes inherent constraints and limits their expressiveness.

\subsection{Chain-of-Thought Reasoning in Continuous Space}

To overcome the inherent limitations of discrete language space in reasoning tasks, recent research has increasingly explored the use of continuous representation spaces for more effective and efficient inference. One pioneering effort in this direction is Coconut~\cite{DBLP:journals/corr/abs-2412-06769}, which introduces the Chain-of-Continuous-Thought paradigm. This approach encodes intermediate reasoning steps as continuous latent vectors, allowing for smooth and information-preserving reasoning trajectories. Building upon this idea, CCoT~\cite{DBLP:journals/corr/abs-2412-13171} proposes a Compressed Chain-of-Thought framework, which generates dense, content-rich continuous representations which is referred to as ``contemplation tokens''. Extending these innovations to multi-modal tasks, Heima~\cite{DBLP:journals/corr/abs-2501-19201} further refines the paradigm by encoding the entire reasoning process into a single continuous vector for multi-modal reasoning. Most recently, SoftCoT~\cite{xu2025softcot} advances this line of work by adapting continuous-space chain-of-thought reasoning to state-of-the-art LLM architectures and mitigates the catastrophic forgetting problem for LLMs with good zero-shot CoT performance.

Despite the promising advances in continuous-space-based chain-of-thought (CoT) reasoning, two major limitations persist in existing approaches. First, none of the current methods incorporate test-time scaling, a widely adopted technique in discrete CoT reasoning for enhancing performance through computational budget expansion during inference. The absence of such scaling mechanisms constrains the effectiveness of continuous-space reasoning on complex downstream tasks. Second, these methods face inherent scalability challenges due to the nature of continuous representations. In discrete token space, multiple diverse reasoning trajectories can be easily obtained via sampling (e.g., temperature sampling), enabling test-time ensembles such as self-consistency. However, in continuous latent space, the representation is deterministic and fixed for a given input, making it non-trivial to generate diverse reasoning paths or multiple hypotheses. This fundamental limitation hinders the scalability of continuous reasoning techniques, especially under settings where diversity and robustness are critical.

These two core limitations motivate the central research question of this work: {\bf How can we enable scalable test-time reasoning in continuous latent space?}

\begin{figure}[t]
  \centering
  \includegraphics[width=0.85\textwidth]{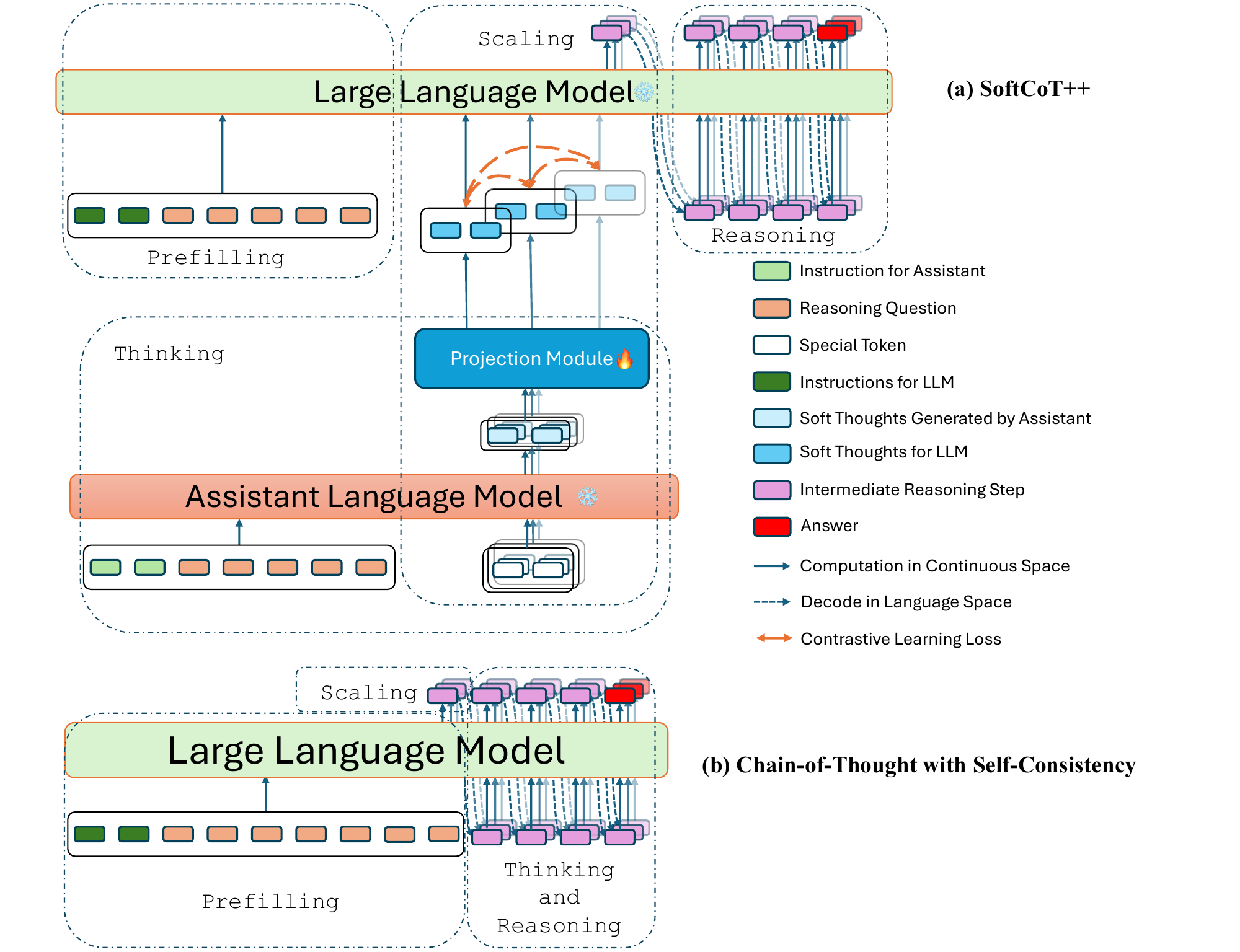}
\caption{A comparison of SoftCoT++ and Chain-of-Thought with Self-Consistency.}\label{fig:softcot++-methodology-overview}
\end{figure}

\section{Methodology}

\subsection{Problem Definition and Notations}
Given a task-specific instruction $\cI=[i_1,i_2,\cdots,i_{|\cI|}]$ and an input query $\cQ=[q_1,q_2,\cdots,q_{|\cQ|}]$, we formalize the problem-solving process of an LLM in three auto-regressive stages:
(1) \textbf{Thinking}. Generate a sequence of thinking steps $\cT=[t_1,t_2,\cdots,t_{|\cT|}]$ based on the input query;
(2) \textbf{Reasoning}. Produce explicit  rationales $\cR=[r_1,r_2,\cdots,r_{|\cR|}]$ based on the query and thinking steps, providing an interpretable reasoning path;
(3) \textbf{Answer Generation}. Output the final answer $\cA=[a_1,a_2,\cdots,a_{|\cA|}]$ conditioned on $\cQ, \cT$ and $\cR$. The generation process can be described as:
\begin{align}
  \label{eq:softcot++-cot-def}
  t_{i+1}&=\mathrm{LLM}\Big([\cI;\cQ;\cT_{\le i}]\Big),&\;\;\;\;//\text{Thinking Process}\\\nonumber
  r_{j+1}&=\mathrm{LLM}\Big([\cI;\cQ;\cT;\cR_{\le j}]\Big),&\;\;\;\;//\text{Reasoning Process}\\\nonumber
  a_{k+1}&=\mathrm{LLM}\Big([\cI;\cQ;\cT;\cR;\cA_{\le k}]\Big),&\;\;\;\;//\text{Answer Generation}\nonumber
\end{align}
where $\mathrm{LLM}(\cdot)$ indicates a large language model, and $[\cdot;\cdot]$ indicates the concatenation of input sequence.
Notably, classical CoT methods~\cite{DBLP:conf/iclr/0001Z0S23,DBLP:conf/iclr/ZhouSHWS0SCBLC23,DBLP:conf/nips/YaoYZS00N23} generate the entire thinking and reasoning steps altogether $\cP =\cT \cup \cR$ using discrete tokens, constraining every step in $\cP$ to lie within the model’s vocabulary space $\cV$.

\subsection{SoftCoT}

Soft Chain-of-Thought (SoftCoT)~\cite{xu2025softcot} introduces a new reasoning paradigm that enhances LLM performance by incorporating continuous latent thoughts. Unlike traditional CoT methods that explicitly generate discrete thinking steps, SoftCoT employs an assistant model to produce latent soft thought tokens. These continuous representations serve as implicit cues, steering the subsequent reasoning chain and boosting the answer accuracy:
\begin{align}
  \label{eq:softcot++-softcot-def}
  \bh^{\mathrm{assist}}&=\mathrm{Assistant}\Big([\cI_{\mathrm{assist}};\cQ;\cS_{1:L}]\Big),\\\nonumber
 \cT_{\mathrm{soft}}&=f_{\theta}\Big(\bh^{\mathrm{assist}}_{|\cI|+|\cQ|+1:|\cI|+|\cQ|+L}\Big),&\;\;\;\;//\text{Thinking Process}\\\nonumber
   \cR_{\mathrm{SoftCoT}}&=\mathrm{LLM}\Big([\cI_{\mathrm{LLM}};\cQ;\cT_{\mathrm{soft}}]\Big),&\;\;\;\;//\text{Reasoning Process}\nonumber
\end{align}
where the assistant model $\mathrm{Assistant}(\cdot)$ receives a task-specific instruction $\cI_{\mathrm{assist}}$, the query $\cQ$, and a placeholder string $\cS_{1:L}$ consisting of special tokens like \texttt{[UNK]} for aggragating $L$ soft thought tokens. It returns hidden states where the last $L$ vectors are taken as the input to the projection module $f_{\theta}(\cdot)$ that maps the representation from assistant model to reasoning model. $\cT_{\mathrm{soft}}=\{h_1,h_2,\dots,h_L|h_i\in\mathbb{R}^d \}$ is the soft thought that replace $\cT$ in Eq~\eqref{eq:softcot++-cot-def}, where $d$ is the dimension of the latent space. In SoftCoT, both assistant model as well as the large reasoning model are fixed, and only trains the parameters in the projection module.

\subsection{Chain-of-Thought Scaling}
\label{sec:softcot++-methodology-sc}

{\bf Definition 1.} {We define the composite function \it $f=a\circ b\circ c$ as a general scaling framework for CoT, where $a$ \textbf{prefills} the input, $b$ is a \textbf{scaling} function that launches $N$ independent reasoning paths, and $c$ is a \textbf{generation} function that completes every path and returns its answer.}

Take chain-of-thought with self-consistency (CoT-SC) as an example, $a$ refers to the initial stage when an LLM encodes $(\cI, \cQ)$ and computes the next-token distribution $P_{\text{LLM}}(x\mid\cI, \cQ)$. $b$ samples $N$ independent reasoning paths $\mathcal P=\{\mathcal P_1,\ldots,\mathcal P_N\}\stackrel{\text{i.i.d.}}{\sim}P_{\text{LLM}}(x\mid\cI, \cQ)$, which are completed by $c$ (the LLM itself), yielding answers $\hat{\mathcal A}_1,\ldots,\hat{\mathcal A}_N$, and output the majority vote. An additional theoretical analysis of when CoT-SC improves CoT is presented in Appendix \ref{appendix:self-consistency}. Notably, $P_{\text{LLM}}(x\mid\cI, \cQ)$ is the distribution from which the $N$ discrete CoT paths are sampled.

\paragraph{Scaling Strategies for SoftCoT.}
SoftCoT follows the same framework but changes what the prefilling step $a$ produces: a latent \emph{soft-thought} $\mathcal T_{\text{soft}} \in \mathbb R^{L\times d}$. There are two reasoning stages in SoftCoT, each of which can support test-time scaling independently:

\begin{itemize}

\item \textbf{Scaling the reasoning stage}:
The soft thoughts $\mathcal{T}_{\text{soft}}$ remain deterministic and are followed by the reasoning stage in discrete space. Since reasoning occurs in a discrete token space, existing scaling methods can be applied directly to this stage. In the baseline model SoftCoT-SC, we adopt the widely used self-consistency approach. Thus, $P_{\text{LLM}}(x\mid\cI, \cQ, \cT_{\text{soft}})$ is the distribution for sampling the $N$ discrete reasoning paths under SoftCoT-SC.

\item \textbf{Scaling the thinking stage}:
Attempting to diversify soft-thought construction. However, because the latent representation $\mathcal{T}_{\text{soft}}$ is deterministic for a given input $(\mathcal{I}, \mathcal{Q})$, standard sampling is not feasible. Thus, ensuring diverse exploration in latent-space reasoning remains a primary challenge. The focus of this paper is to enable scaling in the thinking stage by simulating the multi-path sampling process in continuous space.
\end{itemize}

\begin{figure}[h!]
\centering

\subfloat[Qwen3-8B]{
    \begin{tikzpicture}
    \begin{axis}[
        xlabel={Number of Reasoning Chains},
        ylabel={Accuracy (\%)},
        xmin=0, xmax=11,
        ymin=91.5, ymax=93.5,
        legend pos=south east,
        grid=both,
        width=5cm,
        height=5cm
    ]

    \addplot[
        color=red,
        mark=triangle*,
        thick,
    ] coordinates {
        (1,91.86) (3,92.48) (5, 93.03) (10, 93.10)
    };
    \addlegendentry{SoftCoT-P}

    \addplot[
        color=blue,
        mark=square*,
        thick,
    ] coordinates {
        (1, 91.86) (3, 92.57)  (5, 93.10) (10, 93.18)
    };
    \addlegendentry{SoftCoT-SC}

    \end{axis}
    \end{tikzpicture}
}
\subfloat[LLaMA-3.1-8B-Instruct]{
    \begin{tikzpicture}
    \begin{axis}[
        xlabel={Number of Reasoning Chains},
        ylabel={Accuracy (\%)},
        xmin=0, xmax=11,
        ymin=80.5, ymax=91,
        legend pos=south east,
        grid=both,
        width=5cm,
        height=5cm
    ]
    \addplot[
        color=red,
        mark=triangle,
        thick,
    ] coordinates {
      (1,81.03) (3, 84.91) (5, 89.08) (10,90.58)
    };
    \addlegendentry{SoftCoT-P}

    \addplot[
        color=blue,
        mark=square,
        thick,
    ] coordinates {
        (1,81.03) (3, 84.99) (5, 89.46) (10,90.63)
    };
    \addlegendentry{SoftCoT-SC}

    \end{axis}
    \end{tikzpicture}
}

\caption{Comparison of SoftCoT-P and SoftCoT-SC on GSM8K.\label{fig:softcot++-methodology-pilot-exp}}
\end{figure}
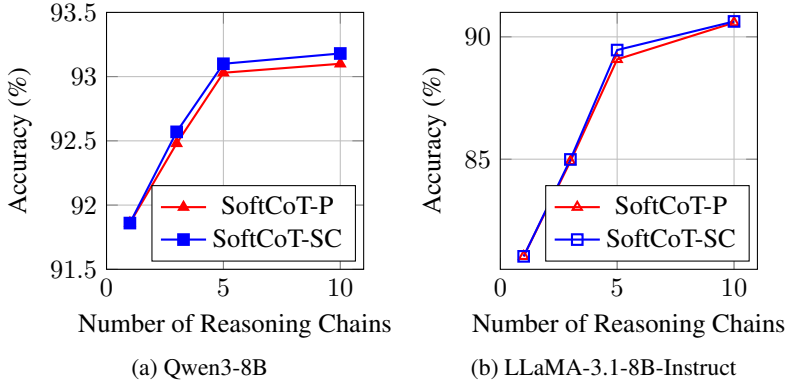

\subsection{Pilot Experiments for Scaling SoftCoT in Thinking Stage}
\label{sec:softcot++-methodology-pilot-experiment}

Let $G_{\phi}= \text{Assistant}\circ f_\theta$ represent the soft-thought generator: $\cT_{\mathrm{soft}}=G(\cI,\cQ, \cS) \in\mathbb R^{d}$. The primary challenge in scaling SoftCoT is the lack of an explicit, sampleable distribution of soft thoughts. To enable this, we make the following assumption.

{\bf Assumption 1.} {\it There exists a \emph{smooth, differentiable} density $P_G(t\mid\cI, \cQ)$, such that the deterministic output $\cT_{\text{soft}}$ may be regarded as a single sample from this density: $\cT_{\text{soft}}\sim P_G(t\mid\cI, \cQ)$.}

{\bf Lemma 1.} {If $\delta$ is sufficiently small, then $\cT_{\text{soft}}+ \delta$ remains in a high-probability region of $P_G$. }

The proof of Lemma 1 is presented in \ref{appendix:proof-of-lemma}.
In other words, according to Lemma 1, {\bf the perturbed sample remains approximately in the same distribution} if the perturbation is small enough. Inspired by this conclusion, it is feasible to add a small perturbation term to the sampled $\cT_{\mathrm{soft}}$, which approximately follows the distribution of soft thoughts:
\begin{align}
  \cT^i_{\mathrm{soft}}=\cT_{\mathrm{soft}}+\delta_i,
\end{align}
where $i$ indicates the $i$-th reasoning chain, and all $\delta_i \to 0$. For convenience, we use ``SoftCoT-P'' (SoftCoT-Perturbed) to mark this scaling method. Compared with SoftCoT-SC, where diversity is injected in explicit token-level sampling, SoftCoT-P injects diversity in the latent space.

Figure~\ref{fig:softcot++-methodology-pilot-exp} shows their performance comparison. We notice that SoftCoT-P has a similar performance compared with SoftCoT-SC, which empirically demonstrates the feasibility of sampling from the estimated latent space distribution. Nevertheless, it does not outperform SoftCoT-SC, which we hypothesize that the small perturbations explore only a narrow neighbourhood of the true density, limiting the diversity of the generated reasoning paths.

\subsection{SoftCoT++}
\label{sec:softcot++-methodology-softcot++}

Let $P=P_{G}(t|\cI,\cQ)$ be the true distribution we need to approximate. As discussed in \S~\ref{sec:softcot++-methodology-pilot-experiment}, a more precise estimation of the soft thought distribution will lead to a better scaling performance. Thus, our goal is to obtain a better representation distribution estimation than SoftCoT-P.

{\bf Definition 2.} {\it Let $\{\delta_i\}_{i=1}^n$ be a set of small perturbations. For each $i$, we define the perturbed soft thought representation as $\cT_{\mathrm{p}}^i = \cT_{\mathrm{soft}} + \delta_i$. The distribution $Q_1$ is the empirical distribution estimated from the set of perturbed samples $\{\cT_{\mathrm{p}}^i\}_{i=1}^n$.}

{\bf Definition 3.} {\it Let $\cT^{\mathrm{scale}}_{\mathrm{soft}}=\{\cT^i_{\mathrm{soft}}\}_{i=1}^n$ be a set of representations sampled from $P_{G}(t|\cI,\cQ)$. The distribution $Q_2$ is then estimated from the $\cT^{\mathrm{scale}}_{\mathrm{soft}}$.}

Based on the definitions, our goal is to {\bf find a distribution $Q_2$ where $\mathrm{KL}(P||Q_2)<\mathrm{KL}(P||Q_1)$}, meaning that $Q_2$ provides a closer approximation to the true distribution $P$ than $Q_1$. Under a mild assumption that $\mathrm{Var}[P] > 0$ and that $\delta_i < \mathrm{Var}[P]$, we have

{\bf Lemma 2.} {\it The candidate distribution $Q_2$ is better than $Q_1$ to describe $P$, if $\mathrm{Var}[Q_1] < \mathrm{Var}[Q_2] \le \mathrm{Var}[P]$, subjects to $\forall\;\cT^i_{\mathrm{soft}}\sim P$.}

The proof of Lemma 2 is shown in \ref{appendix:proof-of-lemma}. Lemma 2 suggests two ways to obtain a better estimation: (1) generate multiple distinct soft thought representations instead of one; (2) encourage higher variance among these soft thought representations.

\paragraph{Diverse Input Sequence.}

Notably, the input of assistant model in Eq~\eqref{eq:softcot++-softcot-def} includes $L$ special \texttt{[UNK]} tokens: $\cS_{1:L}=\texttt{[UNK]}_{1:L}$. Inspiring by the multi-head attention~\cite{DBLP:conf/nips/VaswaniSPUJGKP17} that the structure as well as the computation graph among different head keeps the same but only the initial parameter differs, we replace the special \texttt{[UNK]} token with multiple special \texttt{[INI]} tokens:
\begin{align}
  \hat{\cS}^{i}_{1:L}&=\texttt{[INI]}^{i}_{1:L},\\\nonumber
  \mathrm{s.t.} \quad \texttt{[INI]}^{i}&\ne \texttt{[INI]}^{j},\;\;\forall i\ne j,\nonumber
\end{align}
where $i$ indicates the $i$-th thinking path, and $\texttt{[INI]}^{i}\in \cV$ indicates the $i$-th special initial token for assistant model. The multiple \texttt{[INI]} tokens enables the assistant model to generate multiple soft thoughts:
\begin{align}
  \bh^{\mathrm{assist-}i}&=\mathrm{Assistant}\Big([\cI_{\mathrm{assist}};\cQ;\hat{\cS}^{i}_{1:L}]\Big),\\\nonumber
  \cT^i_{\mathrm{soft}}&=f_{\theta}\Big(\bh^{\mathrm{assist-}i}_{|\cI|+|\cQ|+1:|\cI|+|\cQ|+L}\Big).
\end{align}

\paragraph{Contrastive Learning Loss.}

As aforementioned, a larger variance is required for a better estimation to the target distribution. Thus, we apply the contrastive learning loss as a regulation term to maximize the distance between different thinking representations, which brings a larger variance:
\begin{align}
  \cL_{\mathrm{cl}}=-\sum_{k=1}^{M}\mathbb{E}\Big[\log \frac{\exp(\cT^k_{\mathrm{soft}}\cdot \cT^k_{\mathrm{soft}})}{\sum_{j}\exp(\cT^k_{\mathrm{soft}}\cdot\cT^j_{\mathrm{soft}})}\Big].
\end{align}

\paragraph{Overall Pipeline.} In summary, SoftCoT++ enables test-time scaling in the thinking stage by introducing different special placeholder tokens that provide diverse input embeddings for the assistant model, which can generate multiple soft thinking thoughts. In the training stage, SoftCoT++ also introduces a contrastive loss as the regulation term to enhance the diversity of different soft thinking thoughts, which facilitates to a better estimation of the latent representation distribution.

\section{Experiments}
\label{sec:softcot++-experiments}

\subsection{Datasets}

Following \citet{xu2025softcot}, we conduct experiments on five benchmark datasets spanning three categories of reasoning: mathematical reasoning, commonsense reasoning, and symbolic reasoning. For mathematical reasoning, we utilize GSM8K~\cite{cobbe2021gsm8k}, ASDiv-Aug~\cite{xu2025softcot}, and AQuA~\cite{DBLP:conf/acl/LingYDB17}. For commonsense reasoning, we use StrategyQA~\cite{DBLP:journals/tacl/GevaKSKRB21}, and for symbolic reasoning, we adopt Date Understanding~\cite{DBLP:journals/tmlr/SrivastavaRRSAF23} from the BIG-bench suite. More details can be found in Appendix~\ref{appendix:dataset-stat}.

\subsection{Implementation Details}

We follow the official implementations of SoftCoT~\cite{xu2025softcot}. All models are trained on a single NVIDIA A100-80G GPU. Only the parameters in the projection is trained for 10 epochs. The learning rate is set as 2e-5, and the number of soft thought tokens $L$ is set as 4. To fully utilize the GPU memory, we set the batch size as 8 or 16, which depends on the GPU memory usage.

\subsection{Baselines}
\label{sec:softcot++-experiments-baselines}

As noted by \citet{xu2025softcot}, state-of-the-art LLMs with approximately 8B parameters have strong zero-shot performance on reasoning tasks. However, fine-tuning these models using standard language modeling objectives on reasoning datasets often leads to performance degradation. Consequently, it is crucial to evaluate models under zero-shot settings. We consider the following baselines:

\noindent {\bf Zero-Shot CoT (SC)}: To assess potential degradation caused by supervised fine-tuning, we employ zero-shot chain-of-thought (CoT) prompting using templates from \citet{DBLP:journals/corr/abs-2409-12183}. Self-consistency is enabled, beginning from the initial thinking step, to enhance performance stability.

\noindent  {\bf Zero-Shot Assist-CoT (SC)}: In this baseline, an assistant model is prompted to generate hard reasoning tokens, which are then used for chain-of-thought prompting. Different to the above, we apply self-consistency starting from the reasoning process.

\noindent {\bf Coconut-SC}~\cite{DBLP:journals/corr/abs-2412-06769}: Coconut introduces reasoning in a continuous latent space by recursively feeding intermediate hidden states as input embeddings. This approach facilitates efficient and flexible reasoning compared to traditional discrete CoT methods. We enable self-consistency beginning from the reasoning process.

\noindent  {\bf SoftCoT-SC}~\cite{xu2025softcot}: SoftCoT employs an assistant model to generate fixed soft thoughts, which are then passed to a larger reasoning model to produce the reasoning chain. This setup serves as a baseline where scaling is applied to the reasoning stage of SoftCoT using Self-Consistency, while there is no scaling in the thinking stage.

\section{Results and Discussions}

\begin{table*}[t!]
\setlength{\tabcolsep}{0pt}
    \centering\small
    \begin{tabular}{l| c c c | c || c | c |c c c c}
    \toprule
    \multirow{2}{*}{Model} & GSM8K & ASDiv-Aug & AQuA & Avg. (Math) & StrategyQA & DU & \multirow{2}{*}{Avg. (All)} \\
    \cline{2-7}
    ~ & \multicolumn{4}{c||}{Mathematical} & Commonsense & Symbolic \\
    \midrule
    {\it LLaMA-3.1-8B-Instruct}\\
    Zero-Shot CoT (SC)
    & 90.36$_{\pm \text{0.40}}$ & 89.23$_{\pm \text{0.17}}$ & 63.23$_{\pm \text{0.86}}$ & 80.94 & 70.13$_{\pm \text{0.47}}$ & 65.76$_{\pm \text{1.54}}$ & 75.74 \\
    Zero-Shot Assist-CoT (SC)
    & 90.43$_{\pm \text{0.69}}$ & 89.48$_{\pm \text{0.36}}$ & 63.62$_{\pm \text{0.99}}$ & 81.18 & 70.48$_{\pm \text{0.68}}$ & 65.84$_{\pm \text{1.93}}$ & 75.97 \\
    Coconut-SC~\cite{DBLP:journals/corr/abs-2412-06769}
    & 87.03$_{\pm \text{0.00}}$ & 88.44$_{\pm \text{0.00}}$ & 61.81$_{\pm \text{0.00}}$ & 79.09 & - & - & - \\
    SoftCoT-SC~\cite{xu2025softcot}
    & 90.63$_{\pm \text{0.39}}$ & 89.75$_{\pm \text{0.29}}$ & 65.51$_{\pm \text{0.72}}$ & 81.96 & 71.14$_{\pm \text{0.10}}$ & 67.36$_{\pm \text{1.12}}$ & 76.88 \\
    \midrule
    {\bf SoftCoT++ (Ours)}
    & {\bf 90.99$_{\pm \text{0.25}}$} & {\bf 90.09$_{\pm \text{0.27}}$} & {\bf 66.85$_{\pm \text{0.58}}$} & {\bf 82.64} & {\bf 71.18$_{\pm \text{0.15}}$} & {\bf 68.72$_{\pm \text{0.91}}$} & {\bf 77.57} \\
    \midrule
    \midrule
    {\it Qwen3-8B}\\
    Zero-Shot CoT (SC)
    & 92.22$_{\pm \text{0.47}}$ & 91.97$_{\pm \text{0.13}}$ & 76.77$_{\pm \text{0.62}}$ & 86.99 & 70.96$_{\pm \text{0.15}}$ & 84.56$_{\pm \text{0.61}}$ & 83.30 \\
    Zero-Shot Assist-CoT (SC)
    & 92.68$_{\pm \text{0.17}}$ & 91.91$_{\pm \text{0.28}}$ & 76.77$_{\pm \text{0.79}}$ & 87.12 & 70.92$_{\pm \text{0.28}}$ & 84.80$_{\pm \text{1.17}}$ & 83.42 \\
    Coconut-SC~\cite{DBLP:journals/corr/abs-2412-06769}
    & 90.37$_{\pm \text{0.00}}$ & 90.37$_{\pm \text{0.00}}$ & 76.38$_{\pm \text{0.00}}$ & 85.71 & - & - & - \\
    SoftCoT-SC~\cite{xu2025softcot}
    & 93.19$_{\pm \text{0.32}}$ & 92.14$_{\pm \text{0.15}}$ & 80.63$_{\pm \text{1.90}}$ & 88.65 & 71.18$_{\pm \text{0.15}}$ & 87.20$_{\pm \text{0.75}}$ & 84.87 \\
    \midrule
    {\bf SoftCoT++ (Ours)}
    & {\bf 93.65$_{\pm \text{0.24}}$} & {\bf 92.41$_{\pm \text{0.13}}$} & {\bf 84.09$_{\pm \text{0.72}}$} & {\bf 90.05} & {\bf 71.22$_{\pm \text{0.18}}$} & {\bf 88.16$_{\pm \text{0.54}}$} & {\bf 85.91} \\
    \bottomrule
\end{tabular}
\caption{Model comparison with baselines for test-time scaling. ``SC'' indicates self-consistency, ``DU'' indicates the Date Understanding~\cite{DBLP:journals/tmlr/SrivastavaRRSAF23} dataset. We report results with 10 chains ($N=10$). For all baseline methods, we scale 10 reasoning chains; for SoftCoT++, we scale 10 thinking chains, respectively. We run for 5 random seeds and report the average accuracy as well as the standard deviation.
\label{table:softcot++-result-comparison}}
\end{table*}

\subsection{Comparison with Baselines}

To evaluate SoftCoT++, we compare its performance against the baselines introduced in \S~\ref{sec:softcot++-experiments-baselines}. The results are summarized in Table~\ref{table:softcot++-result-comparison}:

\noindent (1) \textbf{SoftCoT++ successfully extend SoftCoT with test-time scaling}: SoftCoT++ extends SoftCoT by preserving its continuous-thought formulation and introducing two new mechanisms for test-time scaling in its thinking stage: (i) multiple special input tokens that spawn diverse soft-thought trajectories, and (ii) a contrastive regularizer that maintains diversity while preserving informativeness. As shown in Table 1, SoftCoT++ outperforms all baselines, including SoftCoT-SC, across architectures and tasks, demonstrating the effectiveness of applying test-time scaling in continuous latent-space reasoning. Notably, the reduced standard deviations indicate that scaling soft thoughts does not destabilize predictions\textemdash a crucial property for reliable test-time scaling. We hypothesize that the improved performance stems from the increased likelihood of discovering correct answers due to the greater diversity of sampled representations.

\noindent (2) {\bf SoftCoT++ exhibits consistent performance across architectures and tasks}:
Operating entirely at the representation level, SoftCoT++ requires no architecture-specific modifications or tuning. Despite this, it consistently improves performance across both the LLaMA-3 and Qwen-3 model families, demonstrating its backbone-agnostic design. This generality holds regardless of differences in pretraining corpora, tokenization schemes, or positional encoding strategies. Furthermore, SoftCoT++ achieves robust performance across diverse reasoning tasks\textemdash including mathematical, commonsense, and symbolic reasoning\textemdash highlighting its stability and broad applicability. These results confirm that SoftCoT++ enhances reasoning without requiring model-specific adaptation.

\noindent (3) {\bf SoftCoT++ unlocks the latent potential of LLMs via test-time scaling}:
Empirical results on mathematical reasoning tasks show that under flexible inference budgets, the main bottleneck is inference diversity rather than model capacity. SoftCoT++ addresses this by enabling diverse sampling at the representation level, allowing qualitatively distinct inference paths through the same model. This approach better explores the model's internal reasoning capabilities, leading to higher-quality inferences. However, on StrategyQA, we observed diminishing returns when the number of reasoning chains increases to 100, suggesting the model's capacity for that task is already maximised. This contrast underscores SoftCoT++'s ability to fully exploit LLMs’ representational potential, especially in tasks where reasoning diversity remains untapped.

\subsection{Ablation Study}

As discussed in \S~\ref{sec:softcot++-methodology-softcot++}, we theoretically analyzed the importance of diversity in soft thoughts for effectively scaling SoftCoT. Here, we empirically validate this via an ablation study. For clarity, we refer to the variant of our model trained without the contrastive learning objective as ``SoftCoT+''. As shown in Table~\ref{table:softcot++-experiment-self-consistency}, our findings are summarized as follows:

\noindent (1) {\bf SoftCoT+ benefits from scaling}:
Even without contrastive learning, SoftCoT+ shows improved performance across both LLM architectures when scaled using multiple soft thought representations. This confirms the effectiveness of sampling diverse latent representations via different special initial tokens. More detailed discusssion is shown in Appendix~\ref{appendix:discussion-scaling-comparison}.

\noindent (2) {\bf SoftCoT+ underperforms compared to SoftCoT++}:
Despite these improvements, SoftCoT+ performs only marginally better than SoftCoT-SC and significantly worse than SoftCoT++. This result highlights the critical role of contrastive learning in promoting diversity among soft thoughts. Without it, the potential of test-time scaling remains limited, underscoring that the contrastive objective is indispensable for maximizing the benefits of SoftCoT++.

\subsection{The Synergistic Effect of Scaling in the Thinking and Reasoning Stage}

Notably, scaling in the thinking stage is orthogonal to scaling in the reasoning stage. To empirically investigate this distinction, we design an experiment that scales SoftCoT++ along both axes. Specifically, we first generate 10 diverse soft thought representations via SoftCoT++, and then, for each soft thought, apply self-consistency with 10 reasoning chains. This results in a total of 100 reasoning chains per input. As shown in the column $N=100$ of Table~\ref{table:softcot++-experiment-self-consistency}, we compare this scaled version of SoftCoT++ against baseline methods.

The results clearly demonstrate that {\bf SoftCoT++ is orthogonal to self-consistency}.
On one hand, the performance of SoftCoT++ is further improved when combined with self-consistency, highlighting that thinking-stage and reasoning-stage scaling could be used simultaneously to amplify the overall scaling effect. On the other hand, we observe that the performance gain of SoftCoT+ (which lacks contrastive training) from self-consistency is even greater than that of SoftCoT-SC. This indicates that scaling in the thinking stage introduces external benefits beyond what can be achieved by reasoning-stage
scaling alone. A more detailed discussion can be found at \ref{appendix:discussion-scaling-softcot++}.

\begin{table}[t!]
    \centering\small
    \tabcolsep 2.0pt
    \begin{tabular}{l| c c c| c c c }
    \toprule
    \multirow{2}{*}{Model} & \multicolumn{3}{c|}{LLaMA-3.1-8B-Instruct} & \multicolumn{3}{c}{Qwen3-8B} \\
    \cline{2-7}
    ~ & $N=1$ & $N=10$ & $N=100$ & $N=1$ & $N=10$ & $N=100$\\
    \midrule
    Zero-Shot CoT (SC) & 79.61$_{\pm \text{0.81}}$ & 90.36$_{\pm \text{0.40}}$ & 92.42$_{\pm \text{0.21}}$ & 91.86$_{\pm \text{0.41}}$ & 92.22$_{\pm \text{0.47}}$ & 92.98$_{\pm \text{0.11}}$ \\
    Zero-Shot Assist-CoT (SC) & 80.76$_{\pm \text{1.53}}$ & 90.43$_{\pm \text{0.69}}$ & 92.43$_{\pm \text{0.25}}$ & 91.90$_{\pm \text{0.50}}$ & 92.68$_{\pm \text{0.17}}$ & 93.01$_{\pm \text{0.32}}$ \\
    Coconut-SC~\cite{DBLP:journals/corr/abs-2412-06769} & 76.12$_{\pm \text{0.00}}$ & 87.03$_{\pm \text{0.00}}$ & 91.66$_{\pm \text{0.00}}$ & 87.95$_{\pm \text{0.00}}$ & 90.37$_{\pm \text{0.00}}$ & 91.13$_{\pm \text{0.00}}$ \\
    SoftCoT-SC~\cite{xu2025softcot} & 81.03$_{\pm \text{0.42}}$ & 90.63$_{\pm \text{0.39}}$ & 92.52$_{\pm \text{0.17}}$ & 92.48$_{\pm \text{0.29}}$ & 93.19$_{\pm \text{0.32}}$ & 93.40$_{\pm \text{0.15}}$\\
    \midrule
    {\bf SoftCoT+ (Ours)} & - & 90.67$_{\pm \text{0.24}}$ & 92.63$_{\pm \text{0.20}}$ & - & 93.28$_{\pm \text{0.19}}$ & 93.97$_{\pm \text{0.11}}$ \\
    {\bf SoftCoT++ (Ours)} & - & {\bf 90.99$_{\pm \text{0.27}}$} & {\bf 92.71$_{\pm \text{0.14}}$} & - & {\bf 93.65$_{\pm \text{0.24}}$} & {\bf 94.12$_{\pm \text{0.20}}$}\\
    \bottomrule
\end{tabular}
    \caption{Ablation study results on GSM8K. ``$N$'' indicates the number of reasoning chains. ``-'' indicates that when $N=1$, SoftCoT+ and SoftCoT++ reduce to the original SoftCoT. In column $N=10$, we scale 10 reasoning chains for the baseline methods; 10 thinking chains for SoftCoT+ and SoftCoT++. In column $N=100$, we scale 100 reasoning chains for baseline methods. For SoftCoT+ and SoftCoT++, we evaluate the synergistic effect of scaling both the thinking and reasoning stages: we first scale 10 thinking chains and then scale 10 reasoning chains for each thinking chain by self-consistency, resulting in 100 chains in total.}\label{table:softcot++-experiment-self-consistency}
\end{table}

\subsection{Limitations and Future Work}
\label{sec:softcot++-limitation}

Despite the promising results of SoftCoT++, the exploration of the latent thought distribution remains preliminary. In this work, we focus solely on inference with a fixed 8B-scale model. Extending SoftCoT++ to larger, trainable LLMs opens up several promising research directions. In particular, investigating and understanding how the distribution of soft thoughts evolves during training, and how it interacts with model scale and architecture, is a compelling avenue for future work.

\section{Conclusion}

In this paper, we propose SoftCoT++, an extension of SoftCoT that enables test-time scaling in the continuous latent space of the thinking process. SoftCoT++ generates multiple soft thought representations by introducing diverse special tokens as inputs. To encourage representation diversity, we incorporate a contrastive learning objective, which allows the model to more effectively explore the latent solution space. We support our approach with both theoretical analysis and comprehensive empirical evaluation. Experiments across five reasoning benchmarks and two distinct LLM architectures demonstrate that SoftCoT++ consistently improves performance and exhibits strong robustness across settings.

\bibliographystyle{plainnat}
\bibliography{neurips_2025_arxiv}

\begin{thebibliography}{35}
\providecommand{\natexlab}[1]{#1}
\providecommand{\url}[1]{\texttt{#1}}
\expandafter\ifx\csname urlstyle\endcsname\relax
  \providecommand{\doi}[1]{doi: #1}\else
  \providecommand{\doi}{doi: \begingroup \urlstyle{rm}\Url}\fi

\bibitem[BIG.Bench.authors(2023)]{DBLP:journals/tmlr/SrivastavaRRSAF23}
BIG.Bench.authors.
\newblock Beyond the imitation game: Quantifying and extrapolating the
  capabilities of language models.
\newblock \emph{Trans. Mach. Learn. Res.}, 2023, 2023.
\newblock URL \url{https://openreview.net/forum?id=uyTL5Bvosj}.

\bibitem[Brown et~al.(2024)Brown, Juravsky, Ehrlich, Clark, Le, R{\'e}, and
  Mirhoseini]{brown2024large}
Bradley Brown, Jordan Juravsky, Ryan Ehrlich, Ronald Clark, Quoc~V Le,
  Christopher R{\'e}, and Azalia Mirhoseini.
\newblock Large language monkeys: Scaling inference compute with repeated
  sampling.
\newblock \emph{arXiv preprint arXiv:2407.21787}, 2024.
\newblock URL \url{https://arxiv.org/abs/2407.21787}.

\bibitem[Brown et~al.(2020)Brown, Mann, Ryder, Subbiah, Kaplan, Dhariwal,
  Neelakantan, Shyam, Sastry, Askell, Agarwal, Herbert{-}Voss, Krueger,
  Henighan, Child, Ramesh, Ziegler, Wu, Winter, Hesse, Chen, Sigler, Litwin,
  Gray, Chess, Clark, Berner, McCandlish, Radford, Sutskever, and
  Amodei]{DBLP:conf/nips/BrownMRSKDNSSAA20}
Tom~B. Brown, Benjamin Mann, Nick Ryder, Melanie Subbiah, Jared Kaplan,
  Prafulla Dhariwal, Arvind Neelakantan, Pranav Shyam, Girish Sastry, Amanda
  Askell, Sandhini Agarwal, Ariel Herbert{-}Voss, Gretchen Krueger, Tom
  Henighan, Rewon Child, Aditya Ramesh, Daniel~M. Ziegler, Jeffrey Wu, Clemens
  Winter, Christopher Hesse, Mark Chen, Eric Sigler, Mateusz Litwin, Scott
  Gray, Benjamin Chess, Jack Clark, Christopher Berner, Sam McCandlish, Alec
  Radford, Ilya Sutskever, and Dario Amodei.
\newblock Language models are few-shot learners.
\newblock In \emph{Advances in Neural Information Processing Systems 33: Annual
  Conference on Neural Information Processing Systems 2020, NeurIPS 2020,
  December 6-12, 2020, virtual}, 2020.
\newblock URL
  \url{https://proceedings.neurips.cc/paper/2020/hash/1457c0d6bfcb4967418bfb8ac142f64a-Abstract.html}.

\bibitem[Chen et~al.(2025)Chen, Koenig, and Dilkina]{chen2025iterative}
Weizhe Chen, Sven Koenig, and Bistra Dilkina.
\newblock Iterative deepening sampling for large language models.
\newblock \emph{arXiv preprint arXiv:2502.05449}, 2025.
\newblock URL \url{https://arxiv.org/abs/2502.05449}.

\bibitem[Chen et~al.(2024)Chen, Lin, Sch{\"{a}}rli, and
  Zhou]{DBLP:conf/iclr/ChenLSZ24}
Xinyun Chen, Maxwell Lin, Nathanael Sch{\"{a}}rli, and Denny Zhou.
\newblock Teaching large language models to self-debug.
\newblock In \emph{The Twelfth International Conference on Learning
  Representations, {ICLR} 2024, Vienna, Austria, May 7-11, 2024}.
  OpenReview.net, 2024.
\newblock URL \url{https://openreview.net/forum?id=KuPixIqPiq}.

\bibitem[Cheng and Durme(2024)]{DBLP:journals/corr/abs-2412-13171}
Jeffrey Cheng and Benjamin~Van Durme.
\newblock Compressed chain of thought: Efficient reasoning through dense
  representations.
\newblock \emph{arXiv preprint arXiv:2412.13171}, 2024.
\newblock URL \url{http://arxiv.org/abs/2412.13171}.

\bibitem[Chowdhery et~al.(2023)Chowdhery, Narang, Devlin, Bosma, Mishra,
  Roberts, Barham, Chung, Sutton, Gehrmann, et~al.]{chowdhery2023palm}
Aakanksha Chowdhery, Sharan Narang, Jacob Devlin, Maarten Bosma, Gaurav Mishra,
  Adam Roberts, Paul Barham, Hyung~Won Chung, Charles Sutton, Sebastian
  Gehrmann, et~al.
\newblock {PaLM}: Scaling language modeling with pathways.
\newblock \emph{Journal of Machine Learning Research}, 24\penalty0
  (240):\penalty0 1--113, 2023.
\newblock URL \url{https://dl.acm.org/doi/pdf/10.5555/3648699.3648939}.

\bibitem[Cobbe et~al.(2021)Cobbe, Kosaraju, Bavarian, Chen, Jun, Kaiser,
  Plappert, Tworek, Hilton, Nakano, Hesse, and Schulman]{cobbe2021gsm8k}
Karl Cobbe, Vineet Kosaraju, Mohammad Bavarian, Mark Chen, Heewoo Jun, Lukasz
  Kaiser, Matthias Plappert, Jerry Tworek, Jacob Hilton, Reiichiro Nakano,
  Christopher Hesse, and John Schulman.
\newblock Training verifiers to solve math word problems.
\newblock \emph{arXiv preprint arXiv:2110.14168}, 2021.
\newblock URL \url{http://arxiv.org/abs/2110.14168}.

\bibitem[DeepSeek-AI(2025)]{deepseek2025deepseekr1}
DeepSeek-AI.
\newblock {Deepseek-R1}: Incentivizing reasoning capability in llms via
  reinforcement learning.
\newblock \emph{arXiv preprint arXiv:2501.12948}, 2025.
\newblock URL \url{https://arxiv.org/abs/2501.12948}.

\bibitem[Dubey et~al.(2024)Dubey, Jauhri, Pandey, Kadian, Al-Dahle, Letman,
  Mathur, Schelten, Yang, Fan, et~al.]{dubey2024llama}
Abhimanyu Dubey, Abhinav Jauhri, Abhinav Pandey, Abhishek Kadian, Ahmad
  Al-Dahle, Aiesha Letman, Akhil Mathur, Alan Schelten, Amy Yang, Angela Fan,
  et~al.
\newblock The llama 3 herd of models.
\newblock \emph{arXiv preprint arXiv:2407.21783}, 2024.
\newblock URL \url{http://arxiv.org/abs/2407.21783}.

\bibitem[Gandhi et~al.(2024)Gandhi, Lee, Grand, Liu, Cheng, Sharma, and
  Goodman]{gandhi2024stream}
Kanishk Gandhi, Denise Lee, Gabriel Grand, Muxin Liu, Winson Cheng, Archit
  Sharma, and Noah~D Goodman.
\newblock Stream of search (sos): Learning to search in language.
\newblock \emph{arXiv preprint arXiv:2404.03683}, 2024.
\newblock URL \url{http://arxiv.org/abs/2404.03683}.

\bibitem[Geva et~al.(2021)Geva, Khashabi, Segal, Khot, Roth, and
  Berant]{DBLP:journals/tacl/GevaKSKRB21}
Mor Geva, Daniel Khashabi, Elad Segal, Tushar Khot, Dan Roth, and Jonathan
  Berant.
\newblock Did aristotle use a laptop? {A} question answering benchmark with
  implicit reasoning strategies.
\newblock \emph{Trans. Assoc. Comput. Linguistics}, 9:\penalty0 346--361, 2021.
\newblock \doi{10.1162/TACL\_A\_00370}.
\newblock URL \url{https://doi.org/10.1162/tacl\_a\_00370}.

\bibitem[Hao et~al.(2024)Hao, Sukhbaatar, Su, Li, Hu, Weston, and
  Tian]{DBLP:journals/corr/abs-2412-06769}
Shibo Hao, Sainbayar Sukhbaatar, DiJia Su, Xian Li, Zhiting Hu, Jason Weston,
  and Yuandong Tian.
\newblock Training large language models to reason in a continuous latent
  space.
\newblock \emph{arXiv preprint arXiv:2412.06769}, 2024.
\newblock URL \url{http://arxiv.org/abs/2412.06769}.

\bibitem[Li et~al.(2025)Li, Cao, Cao, Li, Tan, Keutzer, Xing, Gonzalez, and
  Stoica]{li2025s}
Dacheng Li, Shiyi Cao, Chengkun Cao, Xiuyu Li, Shangyin Tan, Kurt Keutzer,
  Jiarong Xing, Joseph~E Gonzalez, and Ion Stoica.
\newblock S*: Test time scaling for code generation.
\newblock \emph{arXiv preprint arXiv:2502.14382}, 2025.
\newblock URL \url{https://arxiv.org/abs/2502.14382}.

\bibitem[Lightman et~al.(2024)Lightman, Kosaraju, Burda, Edwards, Baker, Lee,
  Leike, Schulman, Sutskever, and Cobbe]{BoN}
Hunter Lightman, Vineet Kosaraju, Yuri Burda, Harrison Edwards, Bowen Baker,
  Teddy Lee, Jan Leike, John Schulman, Ilya Sutskever, and Karl Cobbe.
\newblock Let's verify step by step.
\newblock In \emph{The Twelfth International Conference on Learning
  Representations}, 2024.
\newblock URL \url{https://openreview.net/forum?id=v8L0pN6EOi}.

\bibitem[Ling et~al.(2017)Ling, Yogatama, Dyer, and
  Blunsom]{DBLP:conf/acl/LingYDB17}
Wang Ling, Dani Yogatama, Chris Dyer, and Phil Blunsom.
\newblock Program induction by rationale generation: Learning to solve and
  explain algebraic word problems.
\newblock In \emph{Proceedings of the 55th Annual Meeting of the Association
  for Computational Linguistics, {ACL} 2017, Vancouver, Canada, July 30 -
  August 4, Volume 1: Long Papers}, pages 158--167. Association for
  Computational Linguistics, 2017.
\newblock \doi{10.18653/V1/P17-1015}.
\newblock URL \url{https://doi.org/10.18653/v1/P17-1015}.

\bibitem[Liu et~al.(2025)Liu, Li, Lv, Liu, Zhu, Hu, and Sun]{liu2025pearl}
Tianyu Liu, Yun Li, Qitan Lv, Kai Liu, Jianchen Zhu, Winston Hu, and Xiao Sun.
\newblock {PEARL}: Parallel speculative decoding with adaptive draft length.
\newblock In \emph{The Thirteenth International Conference on Learning
  Representations}, 2025.
\newblock URL \url{https://openreview.net/forum?id=QOXrVMiHGK}.

\bibitem[Madaan et~al.(2023)Madaan, Tandon, Gupta, Hallinan, Gao, Wiegreffe,
  Alon, Dziri, Prabhumoye, Yang, Gupta, Majumder, Hermann, Welleck,
  Yazdanbakhsh, and Clark]{DBLP:conf/nips/MadaanTGHGW0DPY23}
Aman Madaan, Niket Tandon, Prakhar Gupta, Skyler Hallinan, Luyu Gao, Sarah
  Wiegreffe, Uri Alon, Nouha Dziri, Shrimai Prabhumoye, Yiming Yang, Shashank
  Gupta, Bodhisattwa~Prasad Majumder, Katherine Hermann, Sean Welleck, Amir
  Yazdanbakhsh, and Peter Clark.
\newblock Self-refine: Iterative refinement with self-feedback.
\newblock In \emph{Advances in Neural Information Processing Systems 36: Annual
  Conference on Neural Information Processing Systems 2023, NeurIPS 2023, New
  Orleans, LA, USA, December 10 - 16, 2023}, 2023.
\newblock URL
  \url{http://papers.nips.cc/paper\_files/paper/2023/hash/91edff07232fb1b55a505a9e9f6c0ff3-Abstract-Conference.html}.

\bibitem[Muennighoff et~al.(2025)Muennighoff, Yang, Shi, Li, Fei-Fei,
  Hajishirzi, Zettlemoyer, Liang, Cand{\`e}s, and Hashimoto]{muennighoff2025s1}
Niklas Muennighoff, Zitong Yang, Weijia Shi, Xiang~Lisa Li, Li~Fei-Fei,
  Hannaneh Hajishirzi, Luke Zettlemoyer, Percy Liang, Emmanuel Cand{\`e}s, and
  Tatsunori Hashimoto.
\newblock s1: Simple test-time scaling.
\newblock \emph{arXiv preprint arXiv:2501.19393}, 2025.
\newblock URL \url{https://arxiv.org/abs/2501.19393}.

\bibitem[OpenAI(2023)]{openai2023gpt4}
OpenAI.
\newblock {GPT-4} technical report.
\newblock \emph{arXiv preprint arXiv:2303.08774}, 2023.
\newblock URL \url{http://arxiv.org/abs/2303.08774}.

\bibitem[{Qwen Team}(2025)]{qwen3}
{Qwen Team}.
\newblock Qwen3, April 2025.
\newblock URL \url{https://qwenlm.github.io/blog/qwen3/}.

\bibitem[Renze(2024)]{renze-2024-effect}
Matthew Renze.
\newblock The effect of sampling temperature on problem solving in large
  language models.
\newblock In \emph{Findings of the Association for Computational Linguistics:
  EMNLP 2024}, pages 7346--7356, Miami, Florida, USA, November 2024.
  Association for Computational Linguistics.
\newblock \doi{10.18653/v1/2024.findings-emnlp.432}.
\newblock URL \url{https://aclanthology.org/2024.findings-emnlp.432/}.

\bibitem[Shen et~al.(2025)Shen, Wang, Shi, Wang, Zhao, and
  Gu]{DBLP:journals/corr/abs-2501-19201}
Xuan Shen, Yizhou Wang, Xiangxi Shi, Yanzhi Wang, Pu~Zhao, and Jiuxiang Gu.
\newblock Efficient reasoning with hidden thinking.
\newblock \emph{arXiv preprint arXiv:2501.19201}, 2025.
\newblock URL \url{http://arxiv.org/abs/2501.19201}.

\bibitem[Snell et~al.(2024)Snell, Lee, Xu, and Kumar]{snell2024scaling}
Charlie Snell, Jaehoon Lee, Kelvin Xu, and Aviral Kumar.
\newblock Scaling llm test-time compute optimally can be more effective than
  scaling model parameters.
\newblock \emph{arXiv preprint arXiv:2408.03314}, 2024.
\newblock URL \url{https://arxiv.org/abs/2408.03314}.

\bibitem[Sprague et~al.(2024)Sprague, Yin, Rodriguez, Jiang, Wadhwa, Singhal,
  Zhao, Ye, Mahowald, and Durrett]{DBLP:journals/corr/abs-2409-12183}
Zayne Sprague, Fangcong Yin, Juan~Diego Rodriguez, Dongwei Jiang, Manya Wadhwa,
  Prasann Singhal, Xinyu Zhao, Xi~Ye, Kyle Mahowald, and Greg Durrett.
\newblock To cot or not to cot? chain-of-thought helps mainly on math and
  symbolic reasoning.
\newblock \emph{arXiv preprint arXiv:2409.12183}, 2024.
\newblock URL \url{http://arxiv.org/abs/2409.12183}.

\bibitem[Vaswani et~al.(2017)Vaswani, Shazeer, Parmar, Uszkoreit, Jones, Gomez,
  Kaiser, and Polosukhin]{DBLP:conf/nips/VaswaniSPUJGKP17}
Ashish Vaswani, Noam Shazeer, Niki Parmar, Jakob Uszkoreit, Llion Jones,
  Aidan~N. Gomez, Lukasz Kaiser, and Illia Polosukhin.
\newblock Attention is all you need.
\newblock In \emph{Advances in Neural Information Processing Systems 30: Annual
  Conference on Neural Information Processing Systems 2017, December 4-9, 2017,
  Long Beach, CA, {USA}}, pages 5998--6008, 2017.
\newblock URL
  \url{https://proceedings.neurips.cc/paper/2017/hash/3f5ee243547dee91fbd053c1c4a845aa-Abstract.html}.

\bibitem[Wang et~al.(2025)Wang, Wang, Athiwaratkun, Zhang, and
  Zou]{DBLP:conf/iclr/WangWAZZ25}
Junlin Wang, Jue Wang, Ben Athiwaratkun, Ce~Zhang, and James Zou.
\newblock Mixture-of-agents enhances large language model capabilities.
\newblock In \emph{The Thirteenth International Conference on Learning
  Representations, {ICLR} 2025, Singapore, April 24-28, 2025}. OpenReview.net,
  2025.
\newblock URL \url{https://openreview.net/forum?id=h0ZfDIrj7T}.

\bibitem[Wang et~al.(2023)Wang, Wei, Schuurmans, Le, Chi, Narang, Chowdhery,
  and Zhou]{DBLP:conf/iclr/0002WSLCNCZ23}
Xuezhi Wang, Jason Wei, Dale Schuurmans, Quoc~V. Le, Ed~H. Chi, Sharan Narang,
  Aakanksha Chowdhery, and Denny Zhou.
\newblock Self-consistency improves chain of thought reasoning in language
  models.
\newblock In \emph{The Eleventh International Conference on Learning
  Representations, {ICLR} 2023, Kigali, Rwanda, May 1-5, 2023}. OpenReview.net,
  2023.
\newblock URL \url{https://openreview.net/forum?id=1PL1NIMMrw}.

\bibitem[Wei et~al.(2022)Wei, Wang, Schuurmans, Bosma, Ichter, Xia, Chi, Le,
  and Zhou]{DBLP:conf/nips/Wei0SBIXCLZ22}
Jason Wei, Xuezhi Wang, Dale Schuurmans, Maarten Bosma, Brian Ichter, Fei Xia,
  Ed~H. Chi, Quoc~V. Le, and Denny Zhou.
\newblock Chain-of-thought prompting elicits reasoning in large language
  models.
\newblock In \emph{Advances in Neural Information Processing Systems 35: Annual
  Conference on Neural Information Processing Systems 2022, NeurIPS 2022, New
  Orleans, LA, USA, November 28 - December 9, 2022}, 2022.
\newblock URL
  \url{http://papers.nips.cc/paper\_files/paper/2022/hash/9d5609613524ecf4f15af0f7b31abca4-Abstract-Conference.html}.

\bibitem[Xu et~al.(2025)Xu, Guo, Zeng, and Miao]{xu2025softcot}
Yige Xu, Xu~Guo, Zhiwei Zeng, and Chunyan Miao.
\newblock {SoftCoT}: Soft chain-of-thought for efficient reasoning with llms.
\newblock \emph{arXiv preprint arXiv:2502.12134}, 2025.
\newblock URL \url{https://arxiv.org/abs/2502.12134}.

\bibitem[Yang et~al.(2024)Yang, Yang, Zhang, Hui, Zheng, Yu, Li, Liu, Huang,
  Wei, Lin, Yang, Tu, Zhang, Yang, Yang, Zhou, Lin, Dang, Lu, Bao, Yang, Yu,
  Li, Xue, Zhang, Zhu, Men, Lin, Li, Xia, Ren, Ren, Fan, Su, Zhang, Wan, Liu,
  Cui, Zhang, and Qiu]{DBLP:journals/corr/abs-2412-15115}
An~Yang, Baosong Yang, Beichen Zhang, Binyuan Hui, Bo~Zheng, Bowen Yu,
  Chengyuan Li, Dayiheng Liu, Fei Huang, Haoran Wei, Huan Lin, Jian Yang,
  Jianhong Tu, Jianwei Zhang, Jianxin Yang, Jiaxi Yang, Jingren Zhou, Junyang
  Lin, Kai Dang, Keming Lu, Keqin Bao, Kexin Yang, Le~Yu, Mei Li, Mingfeng Xue,
  Pei Zhang, Qin Zhu, Rui Men, Runji Lin, Tianhao Li, Tingyu Xia, Xingzhang
  Ren, Xuancheng Ren, Yang Fan, Yang Su, Yichang Zhang, Yu~Wan, Yuqiong Liu,
  Zeyu Cui, Zhenru Zhang, and Zihan Qiu.
\newblock Qwen2.5 technical report.
\newblock \emph{arXiv preprint arXiv:2412.15115}, 2024.
\newblock URL \url{http://arxiv.org/abs/2412.15115}.

\bibitem[Yao et~al.(2023)Yao, Yu, Zhao, Shafran, Griffiths, Cao, and
  Narasimhan]{DBLP:conf/nips/YaoYZS00N23}
Shunyu Yao, Dian Yu, Jeffrey Zhao, Izhak Shafran, Tom Griffiths, Yuan Cao, and
  Karthik Narasimhan.
\newblock Tree of thoughts: Deliberate problem solving with large language
  models.
\newblock In \emph{Advances in Neural Information Processing Systems 36: Annual
  Conference on Neural Information Processing Systems 2023, NeurIPS 2023, New
  Orleans, LA, USA, December 10 - 16, 2023}, 2023.
\newblock URL
  \url{http://papers.nips.cc/paper\_files/paper/2023/hash/271db9922b8d1f4dd7aaef84ed5ac703-Abstract-Conference.html}.

\bibitem[Zhang et~al.(2025)Zhang, Lyu, Sun, Wang, Zhang, Guo, Wang, King, Liu,
  and Ma]{zhang2025and}
Qiyuan Zhang, Fuyuan Lyu, Zexu Sun, Lei Wang, Weixu Zhang, Zhihan Guo, Yufei
  Wang, Irwin King, Xue Liu, and Chen Ma.
\newblock What, how, where, and how well? a survey on test-time scaling in
  large language models.
\newblock \emph{arXiv preprint arXiv:2503.24235}, 2025.
\newblock URL \url{https://arxiv.org/abs/2503.24235}.

\bibitem[Zhang et~al.(2023)Zhang, Zhang, Li, and
  Smola]{DBLP:conf/iclr/0001Z0S23}
Zhuosheng Zhang, Aston Zhang, Mu~Li, and Alex Smola.
\newblock Automatic chain of thought prompting in large language models.
\newblock In \emph{The Eleventh International Conference on Learning
  Representations, {ICLR} 2023, Kigali, Rwanda, May 1-5, 2023}. OpenReview.net,
  2023.
\newblock URL \url{https://openreview.net/forum?id=5NTt8GFjUHkr}.

\bibitem[Zhou et~al.(2023)Zhou, Sch{\"{a}}rli, Hou, Wei, Scales, Wang,
  Schuurmans, Cui, Bousquet, Le, and Chi]{DBLP:conf/iclr/ZhouSHWS0SCBLC23}
Denny Zhou, Nathanael Sch{\"{a}}rli, Le~Hou, Jason Wei, Nathan Scales, Xuezhi
  Wang, Dale Schuurmans, Claire Cui, Olivier Bousquet, Quoc~V. Le, and Ed~H.
  Chi.
\newblock Least-to-most prompting enables complex reasoning in large language
  models.
\newblock In \emph{The Eleventh International Conference on Learning
  Representations, {ICLR} 2023, Kigali, Rwanda, May 1-5, 2023}. OpenReview.net,
  2023.
\newblock URL \url{https://openreview.net/forum?id=WZH7099tgfM}.

\end{thebibliography}


\appendix

\section{Theoretical Analysis}
\subsection{Self-Consistency}\label{appendix:self-consistency}
Given an input $x$ with groundtruth $y$, an LLM $\theta$ generates a reasoning chain $r$ conditioned on $x$. SC enables scaling by sampling a set of $N$ independent reasoning paths $\mathcal{R}_N=\{r_i\}_{i=1}^{N}\stackrel{\text{i.i.d.}}{\sim}P_\theta(r\mid x)$ and maps each path to an answer $\hat{y}_i=g(r_i)$. The final prediction $\tilde{y}$ is obtained by majority vote
\[
\tilde{y} \;=\;\operatorname*{arg\,max}_{y}\sum_{i=1}^{N}\mathbf 1\!\bigl[\hat{y}_i=y\bigr].
\]

Let $Z_i=\mathbf 1[\hat{y}_i = y]$ denote the single-path accuracy, where $Z_i\in\{0, 1\}$ follows the Bernoulli distribution. $p$ denotes the probability that a reasoning chain leads to the correct answer. We have
\[
\Pr\bigl(\tilde{y}=y\mid x\bigr)
    \;=\;
    \sum_{k=\lceil N/2\rceil}^{N}\binom{N}{k}p^{\,k}(1-p)^{N-k},
\]
where $k$ is the number of successes.
Here, self-consistency can improve CoT accuracy only when each individual reasoning sample is better than random guessing ($p > 0.5$), so that correct answers are more likely to dominate the sampled set. Increasing $N$ further amplifies the effect of $p > 0.5$ through majority voting. However, raising $p$, e.g., through better prompting or model training, is as important as, and often cheaper than, simply increasing $N$.

\subsection{Proof of Lemmas}\label{appendix:proof-of-lemma}

{\bf Lemma 1.} {If $\delta$ is sufficiently small, then $\cT_{\text{soft}}+ \delta$ remains in a high-probability region of $P_G$. }

\textbf{\textit{Proof of Lemma 1.}} Using Taylor expansion of the probability density function around $x$:
\begin{align}
  p(x+\delta)=p(x)+\nabla p(x)^{\top}\delta+\frac{1}{2}\delta^{\top}\nabla^2p(x)\delta+\cdots
\end{align}
When $||\delta||\to 0$, the higher-order terms vanish faster than the linear term, and:
\begin{align}
p(x+\delta)\approx p(x)+\cO(||\delta||).
\end{align}
So:
\begin{align}
  \frac{p(x+\delta)}{p(x)}\to 1 \;\;\;\;\mathrm{as}\;||\delta||\to 0.
\end{align}
Hence, $x+\delta \sim P_{\theta}(t|c)$ approximately holds for small $\delta$.\hfill $\square$

{\bf Lemma 2.} {\it The candidate distribution $Q_2$ is better than $Q_1$ to describe $P$, if $\mathrm{Var}[Q_1] < \mathrm{Var}[Q_2] \le \mathrm{Var}[P]$, subjects to $\forall\;\cT^i_{\mathrm{soft}}\sim P$.}

\textbf{\textit{Proof of Lemma 2.}} For convenience, we let $P=\cN(\mu,\Sigma)$, $Q_1=\cN(\hat{\mu_1},\hat{\Sigma_1})$, and $Q_2=\cN(\hat{\mu_2},\hat{\Sigma_2})$. Thus we have $\hat{\Sigma}_1 < \hat{\Sigma}_2 < \Sigma$.

Let $P=\cN(\mu,\Sigma)$, and $Q=\cN(\hat{\mu},\hat{\Sigma})$ be two $d$-dimensional Gussians. Then:
\begin{align}
  \label{eq:softcot++-methodology-kl-1}
  \mathrm{KL}(P||Q)=\frac{1}{2}\Big[\underbrace{\mathrm{tr}(\hat{\Sigma}^{-1}\Sigma)}_{\mathrm{first\;term}}+\underbrace{{(\hat{\mu}-\mu)}^{T}\hat{\Sigma}^{-1}(\hat{\mu}-\mu)}_{\mathrm{second\;term}}-d+\underbrace{\log(\frac{\mathrm{det} \hat{\Sigma}}{\mathrm{det} \Sigma})}_{\mathrm{third\;term}}\Big].
\end{align}
Since $\mathbb{E}[\hat{\mu}_1]=\mathbb{E}[\hat{\mu}_2]\approx \mathbb{E}[\mu]$, the second term on both $\mathrm{KL}(P||Q_1)$ and $\mathrm{KL}(P||Q_2)$ is approximated to 0, which can be ignored.
Let $A=\frac{\hat{\Sigma}}{\sigma^2}\Rightarrow \hat{\Sigma}=\sigma^2A$, then Eq~\eqref{eq:softcot++-methodology-kl-1} can be simplified as:
\begin{align}
  \mathrm{KL}(P||Q)&\approx\frac{1}{2}[\mathrm{tr}(A^{-1})-d+\log\mathrm{det}A],\\\nonumber
  &=\frac{1}{2}(f(A)-d),
\end{align}
where $f(A)=\mathrm{tr}(A^{-1})+\log\mathrm{det}A$. Notably, $f(A)$ is minimized when $A=I_d$, then:
\begin{align}
  \mathrm{tr}(A)=d,\log\mathrm{det}A=0\Rightarrow f(A)=d \Rightarrow \mathrm{KL}(P||Q)\approx0.
\end{align}
So the closer $\hat{\Sigma}$ is to $\sigma^2I_d$, the smaller the KL divergence, and hence the better the approximation. Considering $\hat{\Sigma}_1 < \hat{\Sigma}_2 < \Sigma$, we can conclue that $Q_2$ is better than $Q_1$ to describe $P$.\hfill $\square$

\section{Statistical Details for Datasets}
\label{appendix:dataset-stat}

In this section, we present the statistics for datasets we used in Table~\ref{table:softcot++-experiment-dataset}.

\begin{table*}[t!]
    \centering
    \tabcolsep 0pt
    \begin{tabular}{l|c c c c }
    \toprule
    Dataset & Task Type & Answer Type & \# Train samples & \# Evaluation samples \\
    \midrule
    GSM8K~\cite{cobbe2021gsm8k} & \multirow{3}{*}{Mathematical} & Number & 7,473 & 1,319\\
    ASDiv-Aug~\cite{xu2025softcot} & ~ & Number & 4,183 & 1,038 \\
    AQuA~\cite{DBLP:conf/acl/LingYDB17} & ~ & Option & 97,467 & 254 \\
    \midrule
    StrategyQA~\cite{DBLP:journals/tacl/GevaKSKRB21} & Commonsense & Yes/No & 1,832 & 458 \\
    \midrule
    DU~\cite{DBLP:journals/tmlr/SrivastavaRRSAF23} & Symbolic & Option & - & 250 \\
    \bottomrule
\end{tabular}
    \caption{Summary statistics of five datasets we used. ``-'' indicates that there is no training samples available.}\label{table:softcot++-experiment-dataset}
\end{table*}

\section{Discussion}

\subsection{Discussion of the Comparison for Thinking-Scaling and Reasoning-Scaling}
\label{appendix:discussion-scaling-comparison}

As marked in the caption of Table~\ref{table:softcot++-experiment-self-consistency}, SoftCoT+ and SoftCoT++ is the same as SoftCoT when $N=1$. When we scaling to 10 chains (results present in column $N=10$), we notice that SoftCoT+ and SoftCoT++ that scaling 10 thinking chains obtain a larger performance gain than SoftCoT-SC that scaling 10 reasoning chains. The result suggests that scaling in the thinking process in continuous latent space has more potential than scaling in the reasoning process in discrete token space if we have the same computation budget.

Based on this observation, we further try to adopt scaling to the reasoning chain to explore whether scaling in the thinking chain is orthogonal to scaling in the reasoning chain or not. For fairly comparison, we compare the results under 100 chains, which means, for SoftCoT+ and SoftCoT++, there are 10 thinking chains and 10 reasoning chains for each thinking chain, for other baselines, there are only 1 thinking chain and 100 reasoning chains. The experimental results demonstrate that scaling in the thinking process in continuous latent space is orthogonal to scaling in the reasoning process in discrete token space.

\subsection{Discussion of Scaling SoftCoT++ with More Reasoning Chains}
\label{appendix:discussion-scaling-softcot++}

On one hand, the performance of SoftCoT++ is further enhanced when combined with self-consistency, highlighting the complementary strengths of thinking-chain diversity and reasoning-chain aggregation. This improvement suggests that diverse soft thought representations provide a richer set of initial conditions for downstream reasoning, which, when subjected to self-consistency, lead to more robust and accurate final predictions. The combination of diverse thinking paths and multiple reasoning chains allows the model to better explore the solution space, increasing the likelihood of arriving at correct answers through consensus.

On the other hand, the performance gain observed for SoftCoT+ when combined with self-consistency is even more pronounced than that for SoftCoT-SC. This result further emphasizes that enhancing diversity at the thinking level introduces benefits that are not captured by scaling reasoning chains alone. Specifically, even without contrastive regularization, the injection of multiple soft thoughts through distinct initializations enables SoftCoT+ to explore a broader spectrum of latent representations. When self-consistency is applied on top of this diversity, it amplifies the signal from effective reasoning paths while suppressing noise from suboptimal ones.

\end{document}